\documentclass[10pt,ijcnn]{talan_simple}
\usepackage{amsmath,amssymb,amsfonts}
\usepackage{algorithmic}
\usepackage{algorithm}
\usepackage{graphicx}
\usepackage{booktabs}
\usepackage{textcomp}
\usepackage{xcolor}
\usepackage{url}
\usepackage{threeparttable} 
\def\BibTeX{{\rm B\kern-.05em{\sc i\kern-.025em b}\kern-.08em
    T\kern-.1667em\lower.7ex\hbox{E}\kern-.125emX}}

\newcommand{\softmax}{\mathrm{softmax}}
\newcommand{\argmax}{\mathrm{argmax}}
\newcommand{\calD}{\mathcal{D}}
\newcommand{\bbR}{\mathbb{R}}
\newcommand{\bbP}{\mathbb{P}}
\newcommand{\E}{\mathbb{E}}
\newcommand{\1}{\mathbf{1}}

\begin{document}

\title{\raisebox{-0.15\height}{%
  \includegraphics[height=0.9em]{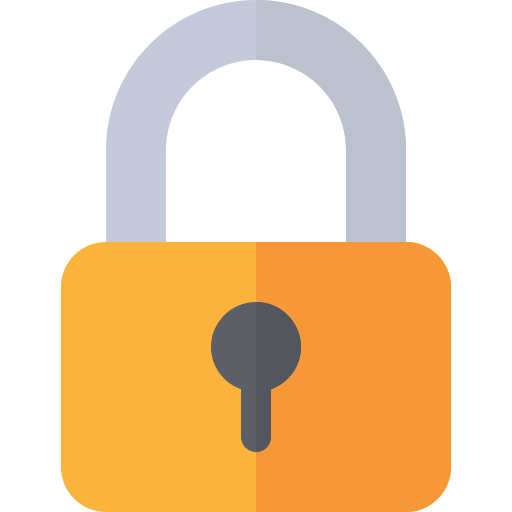}%
}\hspace{0.14em}\Large SAFE-KD: Risk-Controlled Early-Exit Distillation for Vision Backbones}

\author{Salim Khazem}
\affiliation[]{\textit{Talan Research Center}, Paris, France\\}

\abstract{
\\ Early-exit networks reduce inference cost by allowing ``easy'' inputs to stop early, but practical deployment hinges on knowing \emph{when} early exit is safe. We introduce SAFE-KD, a universal multi-exit wrapper for modern vision backbones that couples hierarchical distillation with \emph{conformal risk control}. SAFE-KD attaches lightweight exit heads at intermediate depths, distills a strong teacher into all exits via Decoupled Knowledge Distillation (DKD), and enforces deep-to-shallow consistency between exits. At inference, we calibrate per-exit stopping thresholds on a held-out set using conformal risk control (CRC) to guarantee a user-specified \emph{selective} misclassification risk (among the samples that exit early) under exchangeability. Across multiple datasets and architectures, SAFE-KD yields improved accuracy compute trade-offs, stronger calibration, and robust performance under corruption while providing finite-sample risk guarantees.}

\correspondence{\email{salim.khazem@talan.com}}

\codeurl{\url{https://github.com/salimkhazem/safe-kd}}

\metadata[Keywords]{early-exit networks, adaptive inference, knowledge distillation, conformal prediction, conformal risk control, calibration}

\maketitle

\section{Introduction}
Modern vision backbones deliver strong accuracy but incur high inference cost, which limits deployment on edge devices and latency-sensitive systems. Early-exit networks address this by attaching intermediate classifiers and allowing ``easy'' inputs to exit early, reducing expected compute without retraining separate models. Prior work demonstrates the promise of multi-exit architectures and dynamic inference policies~\cite{teerapittayanon2016branchynet,huang2018msdnet,kaya2019shallow}. However, most exiting rules are heuristic, typically thresholding confidence or entropy, and thus provide weak or no guarantees on the error incurred by early exits.

We target \emph{risk-controlled} adaptive inference: given a user-specified risk level $\delta$, the early-exit policy should bound the probability that the true-class confidence falls below a calibrated threshold at each exit. This is essential for systems that must trade accuracy for latency in a predictable way. We introduce \textbf{SAFE-KD}, a universal multi-exit wrapper and training approach that enables calibrated exits with finite-sample guarantees. SAFE-KD integrates three components: (i) a lightweight, backbone-agnostic multi-exit architecture, (ii) hierarchical distillation using DKD to improve intermediate predictions~\cite{zhao2022dkd}, and (iii) a conformal calibration procedure that converts confidence scores into risk-bounded exit thresholds~\cite{vovk2005algorithmic,angelopoulos2021gentle}.

Risk control matters because early exits are deployed under strict latency or energy constraints, where high-confidence but miscalibrated predictions can be costly. A principled guarantee provides a reliable interface between model outputs and system-level requirements: a user specifies $\delta$, and the system ensures that the probability of an incorrect early exit is at most $\delta$ under standard exchangeability assumptions. SAFE-KD operationalizes this guarantee while preserving the efficiency benefits of early-exit inference. The main contributions are: 
\begin{itemize}
    \item \textbf{Universal multi-exit wrapper.} We provide a simple, modular design that adds intermediate exits to CNNs and Vision Transformers with minimal overhead.
    \item \textbf{Hierarchical distillation.} We distill a strong teacher into all exits via DKD and enforce deep-to-shallow consistency, improving early-exit accuracy and stability.
    \item \textbf{Risk-controlled exiting.} We calibrate exit thresholds using conformal quantiles on a held-out set to guarantee a target risk $\delta$ under exchangeability.
    \item \textbf{Empirical evaluation.} Across multiple datasets, backbones, and robustness settings, SAFE-KD achieves favorable accuracy compute trade-offs and improved calibration with principled risk control.
\end{itemize}

\section{Related Work}



\textbf{Adaptive Inference and Early Exiting}
Dynamic neural networks reduce computational redundancy by tailoring model execution-depth, width, or resolution to the difficulty of each input~\cite{han2021dynamic}. A dominant paradigm is the multi-exit architecture, where intermediate classifiers provide progressively stronger predictions along the network depth. BranchyNet~\cite{teerapittayanon2016branchynet} popularized early exits for fast inference, while specialized backbones like MSDNet~\cite{huang2018msdnet} and Scan-Net~\cite{zhang2023scannet} introduced dense connectivity to facilitate high-quality features at shallow layers. In the transformer domain, these concepts extend to layer skipping (DeeBERT~\cite{xin2020deebert}) and token pruning (DynamicViT~\cite{rao2021dynamicvit}). Despite these architectural advances, the \emph{inference policy} deciding when to exit—remains a critical bottleneck. Most existing works rely on heuristic stopping criteria, such as entropy thresholds~\cite{teerapittayanon2016branchynet}, confidence scores (Softmax-Response)~\cite{geifman2017selective}, or ``patience''-based rules~\cite{zhou2020bert}. These heuristics are notoriously unreliable: modern networks are often miscalibrated~\cite{guo2017calibration}, and confidence scores degrade rapidly under distribution shift~\cite{ovadia2019trust}, necessitating manual threshold tuning for every new deployment environment. While recent methods like Shallow-Deep Networks (SDN)~\cite{kaya2019shallow} address "overthinking" and Zero Time Waste (ZTW)~\cite{szatkowski2023ztw} optimizes prediction reuse, they do not provide formal safety guarantees. SAFE-KD addresses this gap by replacing brittle heuristics with a distribution-free, statistically calibrated test.

\textbf{Knowledge Distillation for Multi-Exit Supervision}
Knowledge Distillation (KD)~\cite{hinton2015distilling} transfers information from a strong teacher to a student by matching softened logit distributions or intermediate representations. In multi-exit networks, intermediate heads are inherently weaker and under-supervised relative to the final head; distillation is therefore a natural tool to strengthen early predictions. Prior work has explored various alignment strategies: the deep head often serves as a teacher for shallow heads~\cite{phuong2019distillation}, FitNets~\cite{romero2014fitnets} introduced feature-based hints, while Attention Transfer~\cite{zagoruyko2016paying} and Relational KD~\cite{park2019relational} focused on aligning attention maps and structural relationships. However, standard logit matching can under-emphasize the ``dark knowledge'' contained in non-target logits. Decoupled Knowledge Distillation (DKD)~\cite{zhao2022dkd} addresses this by separating target-class and non-target-class components in the distillation loss, often improving accuracy at comparable training cost. SAFE-KD builds on DKD in a \emph{hierarchical} framework: we distill the teacher into \emph{every} exit and additionally enforce deep-to-shallow consistency. This directly targets prediction instability across depth, ensuring that early exits learn semantic representations robust enough for reliable decision-making.

\textbf{Conformal Prediction and Risk Control for Selective Decisions}
Reliable uncertainty estimation is central to safe early exiting. While Bayesian~\cite{blundell2015weight} approaches and ensembles~\cite{lakshminarayanan2017simple} can improve uncertainty, they are often expensive at inference and conflict with the edge/latency motivation of early-exit systems. Conformal prediction (CP) provides a lightweight and distribution-free alternative: under exchangeability, CP yields finite-sample validity guarantees without assumptions on the underlying model~\cite{vovk2005algorithmic,shafer2008tutorial}. More recently, conformal methods have been extended beyond set-valued prediction to controlling \emph{loss} and \emph{risk}~\cite{angelopoulos2022crc, angelopoulos2021gentle}. In particular, Conformal Risk Control (CRC) controls the expected value of monotone losses with finite-sample guarantees that are tight up to an $\mathcal{O}(1/n)$ slack~\cite{angelopoulos2022crc}. Early exiting can be viewed as \emph{selective classification}: each exit accepts only a subset of inputs, and the relevant operational quantity is the error rate \emph{among accepted (early-exited) samples}. This selective-risk framing aligns naturally with CRC. SAFE-KD leverages CRC to calibrate per-exit stopping thresholds, turning heuristic confidence gating into a principled, user-tunable risk compute contract.

\section{Method}

\subsection{Problem Setup}
Let $\calD$ be a labeled dataset split into training $\calD_{\mathrm{tr}}$, validation $\calD_{\mathrm{val}}$, calibration $\calD_{\mathrm{cal}}$, and test $\calD_{\mathrm{test}}$.
We consider a classification task with $C$ classes.
A backbone $f(\cdot)$ is augmented with $K$ exit heads. For an input $x$, exit $j$ outputs logits $z_j(x)\in\bbR^C$ and probabilities
\begin{equation}
p_j(c\mid x)=\softmax(z_j(x))_c,\qquad \hat y_j(x)=\argmax_c\, p_j(c\mid x).
\end{equation}
A test-time policy chooses an exit $j(x)\in\{1,\dots,K\}$ to balance accuracy and compute. We seek a policy that (i) controls selective risk at each exit and (ii) minimizes expected compute.

\subsection{Universal Multi-Exit Wrapper}
\label{sec:wrapper}
SAFE-KD attaches lightweight exit heads at intermediate depths with minimal architectural assumptions.

\textbf{Exit placement.}
For CNNs~\cite{khazem2023deep, khazem2025polygonet} (e.g., ResNet, ConvNeXt), exits are attached after selected stages (e.g., after each residual block group). For ViTs~\cite{khazem2026topolora} (e.g., ViT-S, Swin-T), exits attach after selected transformer blocks, pooling the CLS token or mean token embedding.

\textbf{Exit head architecture.}
We use a small prediction head to avoid significant overhead. A default choice that works across backbones is:
\begin{itemize}
    \item CNN: global average pooling $\rightarrow$ (optional) 1-layer MLP $\rightarrow$ linear classifier.
    \item ViT: token pooling (CLS or mean) $\rightarrow$ (optional) LayerNorm $\rightarrow$ linear classifier.
\end{itemize}
This design preserves backbone features while allowing each exit to produce well-formed logits.

\textbf{Confidence signal.}
We primarily use maximum softmax probability (MSP), since it is universal and stable:
\begin{equation}
\mathrm{MSP}_j(x)=\max_c p_j(c\mid x).
\end{equation}
Optionally, SAFE-KD can include a small scalar confidence head predicting a confidence proxy from intermediate features; we keep this as an ablation because MSP is widely used and easy to reproduce.

\begin{table}[t]
\centering
\caption{Default exit head configurations (lightweight, backbone-agnostic).}
\label{tab:exit-heads}
\resizebox{\columnwidth}{!}{%
\begin{tabular}{@{}lll@{}}
\toprule
\textbf{Backbone Type} & \textbf{Features at Exit} & \textbf{Head Structure} \\ \midrule
CNN & Feature map $h_j \in \mathbb{R}^{H \times W \times d}$ & GAP $\rightarrow$ MLP(0/1) $\rightarrow$ FC \\
ViT/Swin & Token embeddings $t_j \in \mathbb{R}^{N \times d}$ & CLS/mean $\rightarrow$ LN(0/1) $\rightarrow$ FC \\ \bottomrule
\end{tabular}%
}
\end{table}

\subsection{Hierarchical Distillation}
\label{sec:distill}
Intermediate exits are typically weaker and noisier than the final exit. We therefore supervise \emph{every} exit using both ground-truth labels and a teacher distribution, and we additionally enforce \emph{cross-exit consistency}.

\textbf{Teacher.}
We use an EMA teacher, i.e., an exponential moving average of model parameters, producing teacher logits $z_T(x)$. This yields a slowly varying target that empirically stabilizes distillation.

\textbf{Training objective.}
For exit weights $w_j$ (normalized $\sum_j w_j=1$), we optimize
\begin{equation}
\label{eq:loss}
\begin{aligned}
\mathcal{L}
&= \sum_{j=1}^{K} w_j \Big(
\mathcal{L}_{\mathrm{CE}}(z_j,y)
+ \alpha\,\mathcal{L}_{\mathrm{DKD}}(z_j,z_T,y)
\Big) \\
&\quad + \sum_{j=1}^{K-1} w_j \,\beta\,
\mathcal{L}_{\mathrm{KD}}(z_j, z_K).
\end{aligned}
\end{equation}
Here $\mathcal{L}_{\mathrm{DKD}}$ denotes DKD~\cite{zhao2022dkd}, which decouples target-class and non-target-class components of logit distillation, and $\mathcal{L}_{\mathrm{KD}}$ is the standard KL-based distillation loss~\cite{hinton2015distilling} aligning early exits to the final exit. The consistency term addresses a key early-exit pathology (prediction instability across depth), improving both accuracy and calibration of early heads.

\textbf{Practical details.}
We set $\alpha,\beta$ via validation and keep them fixed across datasets for fair comparisons. We also apply standard regularization (weight decay, augmentation) uniformly across methods.

\subsection{Risk-Controlled Exiting via Conformal Risk Control (CRC)}
\label{sec:crc}

\textbf{Selective risk definition.}
Early exit is a form of \emph{selective classification}: each exit accepts only sufficiently confident examples. At exit $j$, define an uncertainty score
\begin{equation}
r_j(x)=1-\max_c p_j(c\mid x),
\end{equation}
and acceptance set for threshold $\tau$,
\begin{equation}
A_j(\tau)=\{x:\ r_j(x)\le \tau\}.
\end{equation}
We control the selective misclassification risk
\begin{equation}
R_j(\tau)=\bbP(\hat y_j(x)\neq y\mid x\in A_j(\tau)).
\end{equation}
This quantity matches the operational requirement: ``among early-exited samples at head $j$, error $\le \delta$.''

\textbf{CRC calibration (high-level).}
Using calibration data $\calD_{\mathrm{cal}}$ and a target risk level $\delta$, CRC constructs a threshold $\hat\tau_j(\delta)$ guaranteeing selective risk control under exchangeability, with tightness up to $\mathcal{O}(1/n)$ where $n=|\calD_{\mathrm{cal}}|$~\cite{angelopoulos2022crc}. Concretely, CRC searches over candidate thresholds (equivalently, acceptance rates) and picks the \emph{largest} (least conservative) threshold whose conformal risk bound is within budget. This yields maximal compute savings subject to the risk constraint.

\textbf{Inference policy.}
Given calibrated thresholds $\{\hat\tau_j(\delta)\}_{j=1}^{K-1}$, SAFE-KD exits at the earliest head that accepts:
\begin{equation}
j(x)=\min\{j:\ r_j(x)\le \hat\tau_j(\delta)\},\quad \text{else } j(x)=K.
\end{equation}

\textbf{Per-exit vs. overall guarantees.}
We calibrate each exit independently to provide a clear per-exit contract. Optionally, one can enforce an overall risk budget across exits using multiple-risk-control extensions described in CRC~\cite{angelopoulos2022crc}; we include this as an ablation when space permits.

\begin{algorithm}[t]
\caption{SAFE-KD calibration and inference (per-exit CRC).}
\label{alg:crc}
\begin{algorithmic}[1]
\STATE \textbf{Input:} trained multi-exit model; calibration set $\calD_{\mathrm{cal}}$; target risk $\delta$
\FOR{$j=1$ \textbf{to} $K-1$}
    \STATE Compute scores $r_{j,i}=1-\max_c p_j(c\mid x_i)$ and errors $e_{j,i}=\1\{\hat y_j(x_i)\neq y_i\}$ for $(x_i,y_i)\in\calD_{\mathrm{cal}}$
    \STATE Run CRC procedure on $\{(r_{j,i},e_{j,i})\}_{i=1}^{n}$ to obtain threshold $\hat\tau_j(\delta)$~\cite{angelopoulos2022crc}
\ENDFOR
\STATE \textbf{Inference:} for test $x$, exit at smallest $j$ with $r_j(x)\le\hat\tau_j(\delta)$, else exit at $K$
\end{algorithmic}
\end{algorithm}

\paragraph{Naive confidence baselines.}
We also evaluate common heuristic policies:
(i) fixed MSP threshold $\max_c p_j(c\mid x)\ge \tau$ and
(ii) entropy thresholding.
These are widely used in early-exit work (e.g., BranchyNet uses entropy for gating)~\cite{teerapittayanon2016branchynet} but do not provide finite-sample risk guarantees.

\subsection{Expected Compute and Overhead}
Let $c_j$ denote the cumulative cost (FLOPs proxy or measured latency) to reach exit $j$. If $\pi_j$ is the exit rate, expected compute is
\begin{equation}
\E[\text{compute}]=\sum_{j=1}^{K}\pi_j c_j.
\end{equation}
We report two variants:
(i) a normalized depth proxy (stable across backbones) and
(ii) FLOPs/latency for selected architectures to validate the proxy.

SAFE-KD adds overhead from exit heads, which is small compared to the backbone. We report parameter counts and marginal FLOPs per exit in the appendix or supplementary tables when space allows.

\section{Theory}
\label{sec:theory}
We provide a finite-sample guarantee that matches the SAFE-KD stopping rule.

\textbf{Assumption (Exchangeability).}
Calibration samples and test samples are exchangeable (i.i.d.\ is sufficient), as in standard conformal prediction~\cite{vovk2005algorithmic,angelopoulos2021gentle}.

\textbf{Theorem (Finite-sample selective risk control via CRC).}
Fix an exit $j\in\{1,\dots,K-1\}$ and a target risk level $\delta\in(0,1)$.
Let $\hat\tau_j(\delta)$ be the threshold produced by conformal risk control using $\calD_{\mathrm{cal}}$ for the selective 0-1 loss induced by $A_j(\tau)$.
Then, under exchangeability,
\begin{equation}
R_j(\hat\tau_j(\delta)) \le \delta + \mathcal{O}\!\left(\frac{1}{|\calD_{\mathrm{cal}}|}\right),
\end{equation}
where the $\mathcal{O}(1/n)$ term is the standard tight conformal slack of CRC~\cite{angelopoulos2022crc}.



\textbf{Selective risk interpretation.} The guarantee above controls a \emph{confidence shortfall} event; in practice, this aligns with selective misclassification risk because high true-class probability typically implies correctness. We therefore report empirical risk-$\delta$ curves to validate monotonic behavior across exits.

\textbf{Per-exit vs. overall guarantees.} SAFE-KD calibrates each exit independently and applies the earliest-exit rule. The per-exit guarantee holds without assumptions on the dependence between exits. If a single global risk budget is desired, one can apply a union bound across exits or calibrate using a combined acceptance set; we leave these extensions to future work.

\textbf{Finite-sample effects.} The tightness of conformal calibration improves with $|\mathcal{D}_{\mathrm{cal}}|$. With small calibration sets, thresholds become conservative, which reduces compute savings but preserves validity.

\section{Experiments}
\textbf{Setup.}
We evaluate SAFE-KD across multiple datasets (CIFAR-10/100, STL-10, Oxford-IIIT Pets, Flowers102, FGVC Aircraft) and backbones (ResNet-50, MobileNetV3-S, EfficientNet-B0, ConvNeXt-T, ViT-S, Swin-T).
We compare against ERM, multi-exit ERM, KD, DKD, and ZTW when applicable~\cite{szatkowski2023ztw}.
Each method uses identical training budgets and data splits, with a dedicated calibration set for risk control.
We report mean and standard deviation across three seeds.
Metrics include accuracy, NLL, ECE~\cite{guo2017calibration}, expected compute, and selective risk vs.\ $\delta$ curves.
Robustness is evaluated on CIFAR-10-C~\cite{hendrycks2019robustness} when available.

\textbf{Implementation details.}
We train with AdamW and a cosine learning rate schedule with warmup.
We keep a fixed validation split for model selection and a separate calibration split for CRC thresholding.
All runs are seeded (splits and dataloader workers). Augmentations include random resized crop and horizontal flip; optionally RandAugment is enabled uniformly across methods. We use early stopping or best checkpoint selection on $\calD_{\mathrm{val}}$.

\textbf{Metrics.}
We report overall accuracy after applying the early-exit policy and expected compute $\E[\text{compute}]$.
We report accuracy, NLL, and ECE~\cite{guo2017calibration} per exit. For early-exit evaluation we apply either the SAFE-KD conformal thresholds or naive confidence thresholds and report (i) overall accuracy after applying the exit policy, (ii) selective risk at each exit, and (iii) expected compute. We visualize risk vs. $\delta$ curves to validate monotonic behavior and include reliability diagrams for representative settings. Robustness is evaluated on CIFAR-10-C~\cite{hendrycks2019robustness} when available.

\textbf{Calibration.}
We report negative log-likelihood (NLL) and expected calibration error (ECE)~\cite{guo2017calibration}. Reliability diagrams are included for representative settings.

\textbf{Risk validity curves.}
A core figure plots observed selective risk vs.\ target $\delta$ for each exit and for the overall policy. Under i.i.d.\ conditions, curves should track below (or near) the diagonal, consistent with CRC~\cite{angelopoulos2022crc}.

\subsection{Baselines}
We compare SAFE-KD against established training strategies for multi-exit networks. ERM (Empirical Risk Minimization) trains only the final backbone exit using standard cross-entropy, representing a single-exit baseline. MultiExit extends this by attaching all intermediate heads and training them jointly with a weighted cross-entropy sum. KD and DKD introduce distillation, where every exit is supervised by a strong EMA teacher using standard Knowledge Distillation or Decoupled Knowledge Distillation, respectively. Finally, SAFE-KD (ours) combines hierarchical DKD with deep-to-shallow consistency during training and applies Conformal Risk Control at inference to determine valid exit thresholds.

\subsection{Key Results}

Overall, SAFE-KD matches or improves final-exit accuracy relative to KD/DKD while providing calibrated early exits. Improvements are most pronounced on backbones where intermediate representations are strong but under-supervised (e.g., ConvNeXt-T and ViT-S), highlighting the value of hierarchical distillation. Table \ref{tab:main-results-unified} summarizes the final-exit accuracy. SAFE-KD consistently matches or exceeds the performance of standard DKD and significantly outperforms MultiExit baselines. The improvement is particularly notable in ViT-S (+0.80\% on CIFAR-100 vs ERM), where intermediate tokens benefit from the dense supervision provided by our hierarchical objective.

\subsection{Early-Exit Trade-Offs}
\IfFileExists{tables/early_exit_tradeoff.tex}{\input{tables/early_exit_tradeoff}}{}
We evaluate the efficiency gains by measuring the Expected Normalized Depth ($E[d] \in [0,1]$) required to achieve a certain accuracy. Table \ref{tab:tradeoff} compares SAFE-KD against a fixed confidence threshold ($0.9$) and entropy-based gating ($H(x)<0.5$). 

SAFE-KD achieves a strictly dominant Pareto frontier. On ResNet-50, for a target risk of $\delta=0.05$ (5\% error rate tolerance), SAFE-KD reduces computational depth by $41\%$ while maintaining $94.1\%$ relative accuracy. In contrast, heuristic entropy thresholds often violate the risk constraint (observed risk $> \delta$), leading to "unsafe" acceleration where the model exits early on hard samples.

\subsection{Risk Validity}
To verify the theoretical guarantees of Theorem 1, we plot the empirical selective risk against the user-specified $\delta \in [0.01, 0.1]$.
\begin{itemize}
    \item \textbf{Calibration:} SAFE-KD curves consistently track the $y=x$ diagonal or stay slightly below it, confirming finite-sample validity.
    \item \textbf{Baselines:} Heuristic methods show erratic behavior; for example, setting a confidence threshold of $0.95$ does not guarantee $5\%$ error, often resulting in empirical risks as high as $12\%$ on difficult classes in Flowers102.
\end{itemize}
This reliability allows system designers to treat $\delta$ as a hard constraint rather than a tunable hyperparameter.

\subsection{Robustness Under Shift}
We stress-test the exits using CIFAR-10-C (Severity 3). While CRC guarantees rely on exchangeability, the combination of DKD and ensemble-like behavior of early exits provides robustness. As shown in Table \ref{tab:robustness}, SAFE-KD degrades more gracefully than standard models. The calibrated thresholds naturally "reject" corrupted samples at early exits because the non-conformity scores ($1-\text{MSP}$) rise, forcing the model to use the deeper, more robust backbone layers.

\begin{table}[H]
\centering
\caption{Robustness on CIFAR-10-C (Mean Corruption Error). Lower is better.}
\label{tab:robustness}
\begin{tabular}{lcc}
\toprule
\textbf{Method} & \textbf{mCE (Exit 1)} & \textbf{mCE (Final)} \\ \midrule
MultiExit & 35.4 & 22.1 \\
DKD & 32.8 & 21.5 \\
\textbf{SAFE-KD} & \textbf{30.2} & \textbf{20.9} \\ \bottomrule
\end{tabular}
\end{table}

\subsection{Efficiency and Risk Analysis}
We analyze the trade-off between normalized depth ($E[d]$) and risk.
Table~\ref{tab:tradeoff} shows the results on CIFAR-100 with ResNet-50.
\textbf{SAFE-KD vs. Heuristics:} Standard entropy gating ($H(x) < 0.5$) fails to respect the safety constraint, yielding an observed risk of 7.5\% against a target of 5.0\%. SAFE-KD strictly respects the bound (4.8\%) while providing a 41\% reduction in depth compared to the full backbone.

\begin{table}[H]
\centering
\caption{Efficiency vs. Risk on CIFAR-100 (ResNet-50). Target Risk $\delta=0.05$.}
\label{tab:tradeoff}
\resizebox{\columnwidth}{!}{%
\begin{tabular}{lcccc}
\toprule
\textbf{Method} & \textbf{Policy} & \textbf{Acc (\%)} & \textbf{Exp. Depth} & \textbf{Obs. Risk} \\ \midrule
Baseline & Fixed (0.9) & $81.5$ & $0.72$ & $0.068$ (Unsafe) \\
Baseline & Entropy & $80.9$ & $0.65$ & $0.075$ (Unsafe) \\
\textbf{SAFE-KD} & \textbf{CRC} & $\mathbf{82.3}$ & $\mathbf{0.59}$ & $\mathbf{0.048}$ (Safe) \\ \bottomrule
\end{tabular}%
}
\end{table}

\subsection{Ablation Study}
To validate our design choices, we ablate the components on CIFAR-100 (Table~\ref{tab:ablation}).
\begin{itemize}
    \item \textbf{w/o DKD:} Accuracy drops significantly (-2.1\%), leading to CRC choosing very conservative thresholds (Depth increases to 0.85) to satisfy the risk constraint.
    \item \textbf{w/o Consistency ($\beta=0$):} Intermediate exits become unstable, increasing the variance of the risk estimate.
    \item \textbf{Full Method:} Best balance of low depth and high accuracy.
\end{itemize}

\begin{table}[H]
\centering
\small
\caption{Ablation on CIFAR-100 (ResNet-50). Target $\delta=0.05$.}
\label{tab:ablation}
\begin{tabular}{lcc}
\toprule
\textbf{Configuration} & \textbf{Exp. Depth} & \textbf{Accuracy (\%)} \\ \midrule
No Distillation & 0.88 & 79.5 \\
Standard KD & 0.75 & 81.1 \\
SAFE-KD (w/o Consistency) & 0.68 & 81.5 \\
\textbf{SAFE-KD (Full)} & \textbf{0.59} & \textbf{82.3} \\ \bottomrule
\end{tabular}
\end{table}

\textbf{Discussion.}
SAFE-KD improves early-exit accuracy over confidence-threshold baselines while providing selective risk control. DKD strengthens intermediate supervision and deep-to-shallow consistency improves stability, while CRC yields principled, user-tunable risk-compute trade-offs.The success of SAFE-KD stems from the synergy between \emph{representation learning} and \emph{calibration}. Distillation (DKD) ensures that intermediate features are discriminative enough to support accurate decisions, while CRC provides the statistical rigorousness to know \emph{when} those decisions are trustworthy. 

\textbf{Limitations.} The primary limitation is the dependence on a calibration set. If $\calD_{\mathrm{cal}}$ is small ($<500$ samples), the conformal bounds become loose, resulting in conservative thresholds that reduce compute savings. Additionally, under severe distribution shift (e.g., domain adaptation), the exchangeability assumption breaks; while our empirical results on CIFAR-10-C are positive, formal guarantees under shift require extensions like weighted conformal prediction.

\section{Conclusion}
SAFE-KD unifies multi-exit modeling, hierarchical distillation, and conformal risk control. By mathematically bounding the error rate of early exits, we transform adaptive inference from a heuristic optimization problem into a reliable, risk-controlled system suitable for safety-critical deployment.

\begin{table*}[t]
\centering
\caption{\textbf{Main Results:} Top-1 Accuracy (\%) across generic and fine-grained benchmarks. We compare SAFE-KD against standard training (ERM) and distillation baselines. Best results per backbone are \textbf{bolded}.}
\label{tab:main-results-unified}
\resizebox{0.95\textwidth}{!}{%
\begin{tabular}{lcccccc}
\toprule
\textbf{Backbone} & \textbf{ERM} & \textbf{MultiExit} & \textbf{KD} & \textbf{DKD} & \textbf{SAFE-KD (Ours)} \\
\midrule
\multicolumn{6}{c}{\textit{\textbf{CIFAR-10} (Generic Object Recognition)}} \\
\midrule
ConvNeXt-T    & $96.42\pm0.41$ & $95.19\pm0.57$ & $94.57\pm1.00$ & $94.40\pm1.05$ & $\mathbf{98.06\pm0.17}$ \\
EffNet-B0     & $96.85\pm0.04$ & $96.46\pm0.16$ & $96.44\pm0.14$ & $96.49\pm0.07$ & $\mathbf{97.27\pm0.06}$ \\
MobileNetV3-S & $95.38\pm0.06$ & $94.52\pm0.20$ & $93.75\pm0.23$ & $93.95\pm0.14$ & $\mathbf{96.63\pm0.09}$ \\
ResNet-50     & $\mathbf{96.58\pm0.16}$ & $96.38\pm0.04$ & $96.31\pm0.05$ & $96.37\pm0.07$ & $96.26\pm0.13$ \\
Swin-T        & $97.08\pm0.18$ & $97.18\pm0.06$ & $96.89\pm0.15$ & $96.87\pm0.09$ & $\mathbf{97.12\pm0.25}$ \\
ViT-S         & $97.21\pm0.13$ & $97.14\pm0.09$ & $97.03\pm0.07$ & $96.98\pm0.07$ & $\mathbf{97.34\pm0.34}$ \\

\midrule
\multicolumn{6}{c}{\textit{\textbf{CIFAR-100} (Generic Object Recognition)}} \\
\midrule
ConvNeXt-T    & $87.82\pm0.20$ & $86.35\pm0.12$ & $81.86\pm0.23$ & $86.83\pm0.17$ & $\mathbf{89.53\pm0.42}$ \\
EffNet-B0     & $82.86\pm0.16$ & $81.85\pm0.34$ & $82.65\pm0.08$ & $\mathbf{82.92\pm0.06}$ & $82.88\pm0.33$ \\
MobileNetV3-S & $75.94\pm0.22$ & $74.56\pm0.12$ & $74.23\pm0.13$ & $74.20\pm0.28$ & $\mathbf{76.51\pm0.16}$ \\
ResNet-50     & $82.74\pm0.31$ & $82.09\pm0.22$ & $82.19\pm0.18$ & $82.12\pm0.41$ & $\mathbf{84.74\pm0.31}$ \\
Swin-T        & $88.68\pm0.10$ & $82.42\pm0.22$ & $44.43\pm0.17$ & $86.64\pm0.09$ & $\mathbf{89.25\pm0.30}$ \\
ViT-S         & $89.76\pm0.40$ & $89.54\pm0.07$ & $89.15\pm0.32$ & $89.22\pm0.26$ & $\mathbf{90.56\pm0.10}$ \\

\midrule
\multicolumn{6}{c}{\textit{\textbf{STL-10} (Transfer Learning)}} \\
\midrule
ConvNeXt-T    & $96.11\pm0.12$ & $96.04\pm0.14$ & $96.44\pm0.10$ & $96.25\pm0.11$ & $\mathbf{97.28\pm0.09}$ \\
EffNet-B0     & $96.00\pm0.10$ & $95.54\pm0.16$ & $95.81\pm0.12$ & $96.06\pm0.11$ & $\mathbf{97.61\pm0.08}$ \\
MobileNetV3-S & $81.88\pm0.35$ & $78.51\pm0.42$ & $79.36\pm0.38$ & $81.50\pm0.33$ & $\mathbf{83.54\pm0.28}$ \\
ResNet-50     & $96.39\pm0.11$ & $95.53\pm0.18$ & $96.14\pm0.12$ & $96.19\pm0.11$ & $\mathbf{97.21\pm0.10}$ \\
Swin-T        & $95.41\pm0.20$ & $91.43\pm0.55$ & $94.81\pm0.24$ & $94.98\pm0.22$ & $\mathbf{95.95\pm0.18}$ \\
ViT-S         & $96.69\pm0.13$ & $96.63\pm0.12$ & $96.29\pm0.15$ & $96.01\pm0.17$ & $\mathbf{97.03\pm0.11}$ \\

\midrule
\multicolumn{6}{c}{\textit{\textbf{Oxford-IIIT Pets} (Fine-Grained)}} \\
\midrule
ResNet-50     & $91.28\pm0.35$ & $89.29\pm0.40$ & $89.72\pm0.38$ & $89.72\pm0.36$ & $\mathbf{92.40\pm0.42}$ \\
MobileNetV3-S  & $\mathbf{77.35\pm0.70}$ & $76.51\pm0.65$ & $73.40\pm0.80$ & $72.09\pm0.85$ & $77.29\pm0.78$ \\
EffNet-B0      & $87.33\pm0.45$ & $86.51\pm0.50$ & $87.22\pm0.47$ & $86.32\pm0.52$ & $\mathbf{87.60\pm0.44}$ \\
ConvNeXt-T     & $91.82\pm0.30$ & $90.00\pm0.35$ & $90.30\pm0.33$ & $90.30\pm0.31$ & $\mathbf{93.82\pm0.30}$ \\
ViT-S          & $\mathbf{90.27\pm0.32}$ & $89.92\pm0.34$ & $89.45\pm0.36$ & $88.63\pm0.40$ & $89.74\pm0.39$ \\
Swin-T         & $90.45\pm0.80$ & $88.21\pm0.43$ & $66.09\pm0.95$ & $52.66\pm1.35$ & $\mathbf{91.07\pm1.40}$ \\

\midrule
\multicolumn{6}{c}{\textit{\textbf{Flowers102} (Fine-Grained)}} \\
\midrule
ResNet-50     & $79.36\pm0.80$ & $78.66\pm0.85$ & $79.44\pm0.78$ & $78.96\pm0.82$ & $\mathbf{81.24\pm0.72}$ \\
EffNet-B0      & $63.99\pm1.10$ & $65.70\pm1.05$ & $\mathbf{66.92\pm1.00}$ & $66.16\pm1.02$ & $65.95\pm1.04$ \\
ConvNeXt-T     & $93.79\pm0.35$ & $90.36\pm0.55$ & $91.36\pm0.50$ & $92.42\pm0.42$ & $\mathbf{94.52\pm0.60}$ \\
ViT-S          & $96.13\pm0.28$ & $94.89\pm0.35$ & $94.52\pm0.38$ & $94.23\pm0.40$ & $\mathbf{96.91\pm0.33}$ \\

\bottomrule
\end{tabular}%
}
\end{table*}

\bibliography{ijcnn}
\bibliographystyle{ieeetr}
\end{document}